\documentclass[runningheads]{llncs}

\usepackage{xspace}
\newcommand{\eg}{e.g.\@\xspace}
\newcommand{\ie}{i.e.\@\xspace}
\newcommand{\cf}{cf.\@\xspace}
\newcommand{\vs}{vs.\@\xspace}

\usepackage{graphicx}
\usepackage{booktabs}
\usepackage{amsmath}
\usepackage{amssymb}
\usepackage{xcolor}
\usepackage[accsupp]{axessibility}
\usepackage{cite}

\usepackage[breaklinks,colorlinks,citecolor=blue,linkcolor=blue,urlcolor=blue]{hyperref}
\usepackage{orcidlink}
\usepackage[capitalise,nameinlink]{cleveref} 
\crefname{section}{Sec.}{Secs.}\Crefname{section}{Section}{Sections}
\crefname{table}{Tab.}{Tabs.}\Crefname{table}{Table}{Tables}

\newcommand{\sigb}{\sigma^2_{b}}
\newcommand{\sigW}{\sigma^2_{W}}

\begin{document}

\title{Not All Confusion Is Equal: A Source-Aware Uncertainty Diagnosis for Fine-Grained Aircraft Detection}

\titlerunning{Not All Confusion Is Equal}

\author{Hai Huang \and Helmut Mayer}
\authorrunning{H. Huang et al.}
\institute{Chair of Visual Computing, Institute for Applied Computer Science,\\
Universit\"at der Bundeswehr M\"unchen, Germany\\
\email{hai.huang@unibw.de}}

\maketitle

\begin{abstract}
Fine-grained object detectors are commonly evaluated with confusion matrices,
which show \emph{where} the model is confused but not \emph{why}, nor whether
the confusion can be reduced. We argue that confusion can be attributed to
distinct, separable sources, each quantitatively measurable, turning a passive measurement into actionable guidance. We present \ensuremath{A^2E^2}, a diagnostic
tool that decomposes the sources of confusion along two
axes, $\{\textrm{aleatoric}, \textrm{epistemic}\} \times
\{\textrm{within-class}, \textrm{between-class}\}$, giving a $2\times2$
taxonomy that enumerates the source types. Each quadrant is measured by its
own quantity, computed in one of three places (input geometry, output-space
disagreement, and the bias-parameter posterior), so the two epistemic sources
are separated by construction rather than by an empirical correlation.
On fine-grained aircraft detection, the four quadrants become four named
sources with their own remedy verdict: \emph{affinity} (geometric similarity, irreducible from size alone), \emph{heterogeneity} (geometrically heterogeneous sub-variants, pointing to re-labeling rather than more data), \emph{contested} (an insufficiently trained but learnable boundary, improvable),
and \emph{collapsed} (a class starved of data, reducible). After attributing the confusion to a specific reducible source, we apply a targeted
intervention and verify experimentally that it reduces the diagnosed source specifically while leaving the irreducible sources unchanged.
\ensuremath{A^2E^2} thus turns confusion measurement into a concrete,
validatable and actionable ``diagnosis'' in which the same off-diagonal mass can carry opposite causes and opposite remedies. We also state this framework's limits, including which sources are only partially identifiable on this specific dataset and why.

\end{abstract}

\section{Introduction}
\label{sec:intro}
A confusion matrix is the standard tool for analyzing fine-grained
detection. It reports which pairs of classes a model conflates and which
classes it recovers poorly. Yet it does not report why a given confusion
occurs, whether it can be removed, or how. Two classes may be confused for
distinct reasons: (1) they are close to indistinguishable at the available
sensor resolution; (2) the decision boundary is under-trained; or (3) one of
them has too few examples to compete at all. The first is ``irreducible'',
i.e., an intrinsic limit that additional data cannot change; the latter two are
``actionable'' deficiencies that targeted data or re-balancing can address.

Furthermore, in the context of trustworthy and explainable AI, much of the
field operates at one of two levels: a conceptual level of principles and
taxonomies, or a post-hoc level of saliency and attention maps. Less developed
is a quantitative, attributable layer between them, one that turns ``the model
is unreliable here'' into something measured, attributed to a cause, and acted
upon. Uncertainty quantification (UQ) is a natural place for such a layer, but
standard UQ returns a scalar: a single number indicating that the model should
hesitate, without indicating why, whether the hesitation is warranted, or
whether it can be reduced.

We address this gap by decomposing the uncertainty behind confusion into
source-attributed components: decomposing not the confusion matrix itself,
but directly the sources of the uncertainty that produces confusion. We
organize the decomposition along two dimensions: (1) the standard UQ
distinction between aleatoric uncertainty (intrinsic to the data, irreducible)
and epistemic uncertainty (due to limited knowledge, reducible), and (2)
within-class (intra-class variation a single label cannot absorb, on the
diagonal) and between-class (mistaking one class for another, off-diagonal).
The Cartesian product gives a $2\times2$ taxonomy of four quadrants:
heterogeneity and affinity (two \textbf{A}leatoric sources), collapsed and
contested (two \textbf{E}pistemic sources), i.e., \ensuremath{A^2E^2}
(\cref{sec:framework}). Each quadrant is assigned a single quantity, computed
in one of three parts of the pipeline: the input geometry (a Bhattacharyya
overlap of class size distributions for affinity, a size dispersion for
heterogeneity), the output-space ensemble disagreement (per-pair mutual
information), and the posterior width of the classifier's bias term ($\sigb$).
The two binary partitions are exhaustive, so the four quadrants enumerate the
possible source types (Appendix~\ref{app:completeness}). The partition itself is
structural; the four quantities are complementary readings rather than
statistically independent ones, since the two aleatoric quantities share the
input geometry (Appendix~\ref{app:orthogonality}). After attributing a class's
confusion to a specific reducible source, we predict and test a targeted
remedy (\cref{sec:exp}). The limits of the proposed work on this dataset are
reported in \cref{sec:cannot}.

\section{Related Work}
\label{sec:related}

\paragraph{Aleatoric and epistemic uncertainty.}
Separating predictive uncertainty into an aleatoric part (intrinsic,
irreducible) and an epistemic part (reducible with more data or a better
model) is standard~\cite{kendall2017uncertainties, huellermeier2021aleatoric,
gawlikowski2023survey}. In classification, the epistemic part is commonly the
information-theoretic mutual information between predictions and model
parameters (BALD)~\cite{depeweg2018decomposition, houlsby2011bald,
gal2016dropout}, estimated via MC-dropout~\cite{gal2016dropout}, deep
ensembles~\cite{lakshminarayanan2017deepensembles}, or a Laplace approximation
to the posterior~\cite{mackay1992laplace, ritter2018scalablelaplace,
daxberger2021laplace}. These methods return a per-sample scalar or
aleatoric/epistemic pair, and recent work scrutinizes how cleanly the two can
be disentangled at all~\cite{valdenegro2022deeper}. We differ in two ways: we
attribute \emph{class-level} confusion, not individual predictions, to a
within/between $\times$ aleatoric/epistemic grid, and show the epistemic side
is not one phenomenon but two: a contested boundary and a collapsed class that
a single MI value cannot separate, since they live in different model parts.
Closest to this point, Toure and Stephens~\cite{toure2026notjust} decompose MI
into per-class contributions and show that output-space variance is suppressed
for rare classes; we build on the same observation but measure collapse in the
parameter space, which remains informative when a class is never predicted.

\paragraph{Confusion, error detection, and attribution.}
A large body of work flags \emph{which} predictions are likely wrong,
via misclassification and out-of-distribution
detection~\cite{hendrycks2017baseline}, or, in fine-grained recognition, via
confusion matrices, hard-pair mining, and class-similarity analyses. These
tell us where a model fails but not \emph{why}, nor whether intervention
would help. We complement error \emph{detection} with source
\emph{attribution}, assigning each confusion a cause and a remedy verdict.

\paragraph{Fine-grained recognition and remote-sensing aircraft.}
Fine-grained recognition features small inter-class and large intra-class
variation~\cite{wei2022finegrained}, classically addressed by localizing
discriminative parts~\cite{zhang2014partbased}; this is the lineage of our
within-class heterogeneity source, where a single label spans several
sub-variant geometries. Fine-grained aircraft recognition in remote sensing is
driven by benchmarks such as FAIR1M~\cite{sun2022fair1m},
MAR20~\cite{yu2023mar20}, and the Gaofen challenge~\cite{sun2021gaofen} on
which we instantiate the framework, typically with oriented
detectors~\cite{xie2021orientedrcnn, zhou2022mmrotate}: we use this setting not
to advance detection accuracy but because oriented boxes give a clean physical
handle on class size geometry.

\paragraph{Data-centric and actionable uncertainty.}
Closest in spirit is work that turns uncertainty into an action on the data.
Data-IQ~\cite{seedat2022dataiq} uses aleatoric uncertainty to stratify
\emph{examples} into easy, ambiguous, and hard subgroups; classical active
learning~\cite{houlsby2011bald} uses epistemic uncertainty to pick which
examples to label next. We share the premise that decomposed uncertainty should imply an
action, but operate on \emph{confusion sources} rather than individual
examples: each quadrant yields a distinct verdict: add class data, add
boundary data, re-label into sub-variants, or leave alone as irreducible.

\paragraph{Trustworthy and explainable AI.}
Research on trustworthy and explainable AI ranges from high-level desiderata
to post-hoc, instance-level explanations such as saliency and attention
maps~\cite{guidotti2018survey, arrieta2020xai}. Our contribution sits in the
quantitative middle: a measured, source-attributed account linking an
observed failure to a cause and a remedy. We treat trustworthiness as
motivation, not a solved problem, and do not claim mechanistic
interpretability: our quantities read the last classification layer, not the
backbone.

\section{The \ensuremath{A^2E^2} Framework}
\label{sec:framework}

\subsection{Setup}
\label{sec:setup}

We conduct our study on the Gaofen fine-grained aircraft benchmark~\cite{sun2021gaofen}, which has nine aircraft classes plus a catch-all ``other'' category (ten classes in total). As baseline model we employ the widely used Oriented R-CNN~\cite{xie2021orientedrcnn}, which localizes aircraft and classifies their type at the family level. The confusion we decompose is the classification part given the oriented bounding box (OBB), i.e., ``confusion'' here refers to class confusion, not localization error.

Each detection is an OBB with corners
$P_1,\dots,P_4$. We define the body (fuselage) length and wingspan by a fixed
corner convention rather than by sorting, because some aircraft have a wing
span larger than their fuselage length, so a $\max/\min$ convention would
silently swap the two axes. The Gaofen images mix two ground sample distances
(GSD $\approx 0.810$\,m for about $80\%$ of images and $\approx 0.536$\,m for
the rest), so pixel-based sizes are converted to physical size with the known
per-image resolution:
\begin{equation}
\ell^{(\textrm{m})}_{\textrm{body}} = \lVert P_2 - P_1 \rVert, \qquad
\ell^{(\textrm{m})}_{\textrm{wing}} = \lVert P_3 - P_2 \rVert.
\end{equation}

Size statistics and the uncertainty estimates are computed on the
development set (train and validation, $2{,}000$ images and
$10{,}957$ instances in total); the held-out test split of $1000$ images is
kept model-unseen. For the epistemic quantities we place a diagonal
Laplace approximation~\cite{mackay1992laplace, ritter2018scalablelaplace,
daxberger2021laplace} on the final classification layer
(a single linear map from the $1024$-dimensional pooled feature to the class
logits) and draw an ensemble of $M=16$ weight samples. This is a last-layer,
post-hoc construction: it reuses the trained checkpoint and adds no training, so
we read uncertainty only from this layer and make no claim about the backbone.

\subsection{Two dimensions, four quadrants}
\label{sec:quadrants}

\begin{figure}[h!]
\centering
\includegraphics[width=\textwidth]{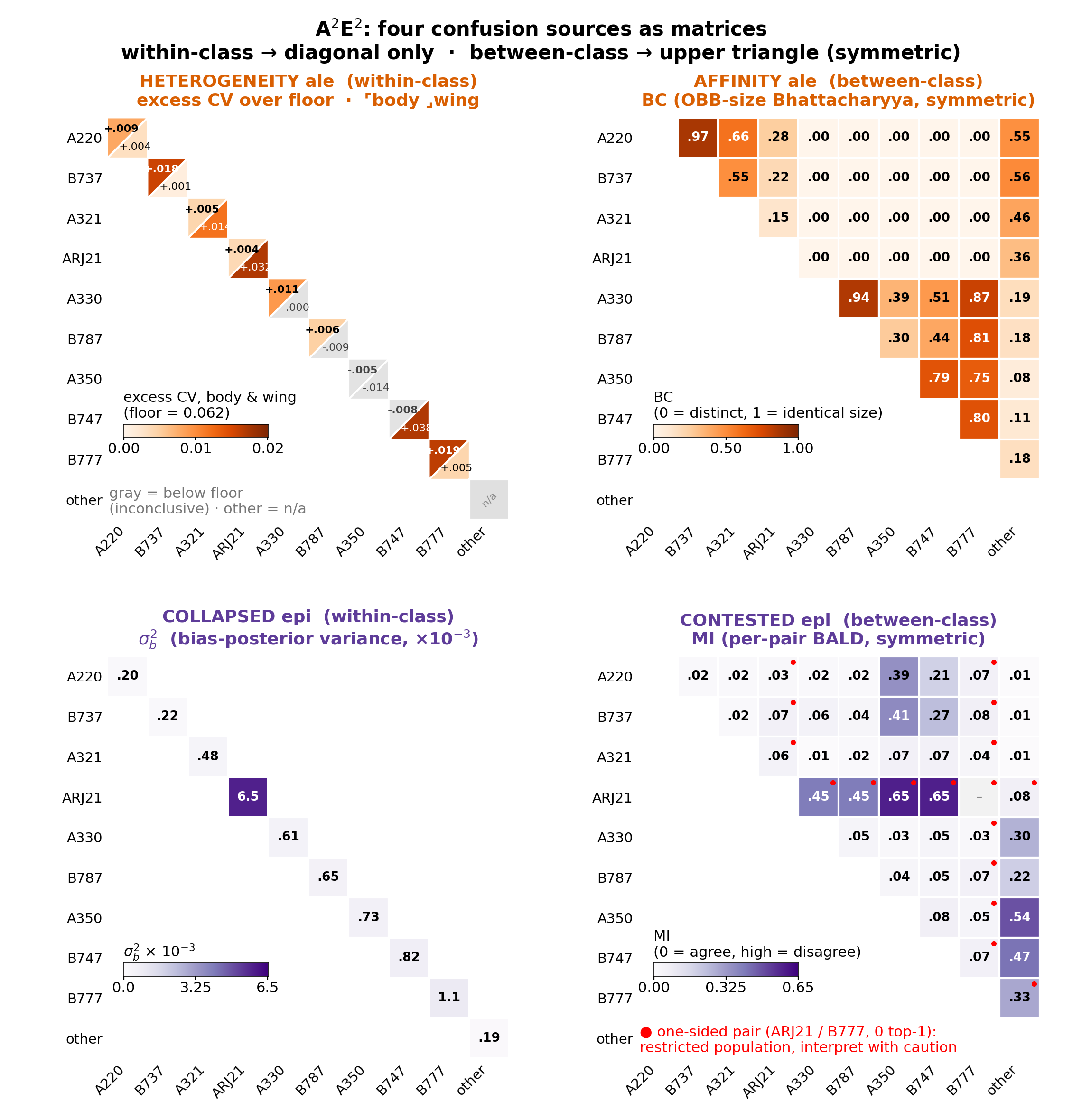}
\caption{Overview of \ensuremath{A^2E^2}: the four confusion sources as matrices over the
ten classes. Within-class sources occupy the diagonal only, between-class
sources the (symmetric) upper triangle. \emph{Affinity} (between-ale) is the
OBB-size Bhattacharyya overlap BC (\cref{eq:bc}); \emph{contested}
(between-epi) is the per-pair mutual information MI (\cref{eq:mi});
\emph{collapsed} (within-epi) is the bias-posterior variance $\sigb$
(\cref{eq:sigb}); \emph{heterogeneity} (within-ale) is the size dispersion (CV)
over a noise floor, split per diagonal cell into body/wing (collapsed classes
shown but confounded by scarcity, \cf{} \cref{fig:cv}). Aleatoric sources
in orange, epistemic in purple. Red dots mark MI pairs involving ARJ21 or
Boeing777 (zero top-1 predictions at baseline), computed on a restricted
population (\cref{sec:metrics}) and not read as contested.}
\label{fig:flagship}
\end{figure}

We split the confusion along two binary dimensions. The first is
the standard uncertainty dichotomy: a source is \emph{aleatoric} (intrinsic to
the data, irreducible) or \emph{epistemic} (due to limited knowledge,
reducible). The second is structural to confusion: a confusion is
\emph{between-class} (one class mistaken for another, a property of a pair) or
\emph{within-class} (intra-class variation a single label cannot absorb, a
property of one class). Their product gives four quadrants, each with its own
name and metric, illustrated in \cref{fig:flagship}.

Two of the quadrants deserve emphasis, since the epistemic side is not a
single phenomenon: a \emph{contested} boundary lives in the output space
(metric: per-pair MI, both classes considered but under-trained), while a
\emph{collapsed} class lives in the bias parameter (metric: $\sigb$, too few
examples to compete). They are different phenomena in different parts of the
model, and one metric is structurally blind to the other: MI is high only
when posterior samples \emph{disagree} about the winner, but a collapsed
class loses \emph{consistently} across samples, so the ensemble agrees and MI
stays near zero even though the class is far from healthy
(Appendix~\ref{app:orthogonality} gives the full argument). On the aleatoric side,
between-class \emph{affinity} (shared physical size, metric BC) is likewise distinct from
within-class \emph{heterogeneity} (sub-variant geometries under one label,
metric: size dispersion over a noise floor, \cref{sec:metrics}).

\subsubsection{What is claimed about completeness and separation.}
The two partitions are exhaustive, so the four quadrants enumerate the
\emph{types} of confusion source rather than a convenient subset
(Appendix~\ref{app:completeness}). The partition into two axes and four quadrants is
structural; the four \emph{quantities}, however, are complementary rather than
independent: the two epistemic quantities live in different parts of the
model (output space \vs{} bias posterior), while the two aleatoric quantities
both read the input geometry (Appendix~\ref{app:orthogonality}).

\subsection{One quantity per quadrant}
\label{sec:metrics}

We now define the quantity assigned to each quadrant, where in the model it
is computed, and its physical and statistical reading.

\subsubsection{Affinity (between-class aleatoric): geometric overlap via Bhattacharyya coefficient (BC).}
For each class we fit a two-dimensional Gaussian to its physical size features
$(\ell^{(\textrm{m})}_{\textrm{body}}, \ell^{(\textrm{m})}_{\textrm{wing}})$
over the development set, obtaining a mean $\mu_a$ and covariance $\Sigma_a$.
For a pair $(a,b)$ we measure the overlap of the two size distributions by:
\begin{equation}
\textrm{BC}(a,b) = \exp\!\big(-D_B(a,b)\big),\quad
D_B = \tfrac{1}{8}(\mu_a-\mu_b)^{\!\top}\Sigma^{-1}(\mu_a-\mu_b)
      + \tfrac{1}{2}\ln\frac{\lvert\Sigma\rvert}{\sqrt{\lvert\Sigma_a\rvert\,\lvert\Sigma_b\rvert}},
\label{eq:bc}
\end{equation}
with $\Sigma=\tfrac{1}{2}(\Sigma_a+\Sigma_b)$. BC lies in $[0,1]$: it is $1$
when the two size distributions coincide and approaches $0$ as they separate.
This quantity is computed entirely in the \emph{input geometry} and never
consults the model. Physically it measures whether two aircraft types are the
same size and shape on the ground, an intrinsic, data-independent property: no
amount of additional training data can make two equal-size airframes
geometrically distinguishable. It therefore measures the between-class
aleatoric source we call \emph{affinity}, irreducible \emph{from size alone}:
high BC means the two types are geometrically close kin in size (appearance
cues beyond the OBB may still separate them).

\subsubsection{Contested (between-class epistemic): per-pair mutual information (MI).}
Using the $M=16$ last-layer posterior samples, we restrict attention to the two
logits of a pair $(a,b)$, renormalize the softmax over those two classes, and
obtain a two-class predictive distribution $p_{ab}^{(m)}$ from each ensemble
member $m$. We then apply the standard information-theoretic
decomposition~\cite{houlsby2011bald, gal2016dropout},
\begin{align}
&C_{\textrm{total}}(a,b) = H\!\Big(\tfrac{1}{M}\textstyle\sum_m p_{ab}^{(m)}\Big),
C_{\textrm{ale}}(a,b) = \tfrac{1}{M}\textstyle\sum_m H\!\big(p_{ab}^{(m)}\big), \\
&\textrm{MI}(a,b) = C_{\textrm{total}}(a,b) - C_{\textrm{ale}}(a,b) \;\ge\; 0,
\label{eq:mi}
\end{align}
where $H$ is the Shannon entropy. MI is the mutual information between the
prediction and the model parameters (BALD): the part of the total two-class
uncertainty coming from the ensemble \emph{disagreeing} about which of $a,b$
wins. It is computed in the model's \emph{output space}. MI is high when the
posterior samples disagree, the signature of an under-trained but learnable
boundary, and near zero when the ensemble agrees, even if it agrees for the
wrong reason. MI is evaluated on detections in which the pair is actually
in play. For 28 pairs both classes have top-1 detections (\emph{valid}
pairs). Two classes, ARJ21 and Boeing777, receive no top-1 predictions at
baseline; their pairs are either recovered from detections whose top-2
contains both classes (\eg{} ``other''--ARJ21, $n=300$), or computed on the
partner class's detections only (\emph{one-sided}), a restricted population
that does not probe a two-class competition and which we do not read as
contested.

\subsubsection{Collapsed (within-class epistemic): bias-posterior variance ($\sigb$).}
The per-class logit variance under the last-layer posterior decomposes into a
weight-side and a bias-side term,
\begin{equation}
\mathrm{Var}_m\big[\textrm{logit}_X\big] = \underbrace{\sum_d z_d^2\,\sigma^2_{W,X,d}}_{\textrm{Weight term, input-dependent}}
 + \underbrace{\sigma^2_{b,X}}_{\textrm{Bias term, input-independent}},
\label{eq:sigb}
\end{equation}
where $z$ is the pooled feature. We take the bias-side term $\sigb \equiv
\sigma^2_{b,X}$, the posterior variance of the classifier's bias for class $X$,
as the collapsed-epistemic quantity, computed in the \emph{bias-parameter
posterior}: the class's input-independent prior tendency, its baseline logit
before any image feature is considered. When a class has too few training
examples, this prior never firms up and its bias posterior stays wide, so the
class is overwhelmed in competition regardless of the input. $\sigb$ is
therefore a per-class, undirected quantity: it measures \emph{that} a class is
collapsing, by how scarce its evidence is, but under a diagonal Laplace
posterior it cannot say \emph{which} competitor it collapses toward (requiring
off-diagonal, cross-class covariance, which the diagonal approximation sets to
zero; see \cref{sec:cannot}). On this dataset $\sigb$ is close to a
deterministic function of the training count (Pearson $r=+0.997$ with $1/n$
across the ten classes, stable over three seeds). This is the expected
behavior of a Laplace posterior on a bias term, and unlike $\sigW$ below,
whose target was geometry rather than scarcity, it is confirmatory rather than
disqualifying; but it also means that, as a ranking, $\sigb$ adds little
beyond per-class counts, so we rest the collapsed diagnosis on the
intervention of \cref{sec:treatment} rather than on the ranking alone.

\subsubsection{Heterogeneity (within-class aleatoric): size dispersion over a noise floor.}
The fourth quadrant captures intra-class geometric diversity: one label
covering several real sub-variants with different airframes (\eg{} A330-200
\vs{} A330-300). A single-variant class, measured through oriented boxes, still
has non-zero size dispersion from measurement noise alone; the signal of
heterogeneity is therefore dispersion \emph{in excess of} that noise. We
quantify it with the class-conditional coefficient of variation
$\textrm{CV}=\sigma/\mu$ of each physical size dimension, referenced to a noise
floor estimated from classes single-variant in this dataset (A321 and A350,
whose dominant variant covers $\sim\!90\%$ of instances). The floor is the mean
CV of these reference classes, $\approx0.062$ for both body and
wing.\footnote{Since the floor is the mean of the two references, each
reference sits within $\pm0.005$ of it in body length ($\pm0.014$ in
wingspan) by construction; ``at the floor'' refers to whichever dimension is
under discussion.} A class
whose CV exceeds the floor carries dispersion beyond measurement noise,
consistent with sub-variant heterogeneity; a class at or below the floor is
inconclusive, i.e., insufficient evidence, not proof of absence. We report CV for
body and wing separately rather than averaged, because a class's variants may
differ in one dimension but not the other (\eg{} A330-200 \vs{} -300 differ in
fuselage length but share a wingspan), and averaging would let the quieter
dimension mask the signal. This quantity is computed in the \emph{input
geometry} (like BC), and, being a property of the size data rather than of any
posterior, it is aleatoric by construction.

A tempting last-layer alternative, the weight-side posterior variance
$\sigW\equiv\sigma^2_{W,X,d}$ from \cref{eq:sigb}, is \emph{not} suitable
here: it tracks training-sample count ($\rho=-0.91$) rather than sub-variant
geometry, \ie{} it is collapsed-epistemic in disguise. We give the evidence
for rejecting it (Appendix~\ref{app:sigmaW}), and several peak-finding
alternatives (Appendix~\ref{app:hetero}); the metric we retain is the
input-geometry CV above.

For each quadrant we select clean ``anchors'', i.e., exemplar classes or pairs
as our study cases, demonstrated in \cref{sec:exp} to both validate the
quantities and assign each a remedy verdict.

\section{Diagnosis with a Remedy Verdict}
\label{sec:exp}

The proposed framework is only meaningful if it provides insight and guides
action against confusion. In this section we study a set of hard cases, and
for each we (1) attribute the confusion to a quadrant and (2) issue a
\emph{remedy verdict}: a statement of whether, and how, the confusion can be
reduced. Four cases, one per quadrant, cover the full range of verdicts
(\cref{tab:cases}): a class whose confusion is reducible and repaired (ARJ21,
collapsed); a pair reducible in principle and supported (A220--A350, contested);
a pair that is irreducible from size, so no amount of data will help (A330--Boeing787,
affinity); and a class also irreducible but for a different reason, calling for
a different, non-data remedy (A330, heterogeneity).

\begin{table}[h!]
\centering
\caption{Four study cases with remedy verdict covering the four quadrants
}
\label{tab:cases}
\resizebox{\textwidth}{!}{%
\begin{tabular}{lllll}
\toprule
\textbf{Case} & \textbf{metric} & \textbf{value} & \textbf{Attributed source} & \textbf{Remedy verdict} \\
\midrule
ARJ21 (class)          & $\sigb$ & rank 1/10 & collapsed (within-epi)   & add class data \emph{(verified, T1)} \\
A220--A350 (pair)      & MI  & $0.387$   & contested (between-epi)  & add boundary data (supported, directional, T2) \\
A330--Boeing787 (pair) & BC  & $0.945$   & affinity (between-ale)   & irreducible from size; more data will not help \\
A330 (class)           & excess-CV & $+0.011$ body & heterogeneity (within-ale) & more data will not help; re-label (initial test null) \\
\bottomrule
\end{tabular}}
\end{table}

Before the individual cases, one contrast makes the point of the whole paper
concrete. A330--Boeing787 and A220--A350 are both confusable pairs, yet sit at
opposite corners of the framework. A330 and Boeing787 are two wide-body
airframes of almost the same physical size: their size distributions overlap
at $\textrm{BC}=0.945$, while their contested uncertainty is low
($\textrm{MI}=0.047$). A220 and A350 are a narrow-body and a wide-body of
clearly different size: their size overlap is essentially zero
($\textrm{BC}\approx0$), while their contested uncertainty is among the
highest of all aircraft pairs ($\textrm{MI}=0.387$; only Boeing737--A350 is
higher, $0.41$), visible in \cref{fig:scatter}
(Appendix~\ref{app:sizegeo}). The
two pairs are confused for opposite reasons: one because the airframes are
genuinely the same size, the other despite being different sizes, and, as
the verdicts below show, call for opposite responses. A geometric overlap does
not imply boundary disagreement, and vice versa; this single contrast is also
the most direct empirical illustration of the affinity/contested separation
argued in Appendix~\ref{app:orthogonality}.

\subsection{ARJ21: collapsed, reducible and verified}
\label{sec:case-arj21}

ARJ21 is the cleanest collapsed class in the dataset: it has (1) the highest
data scarcity, (2) the highest size-difference from other classes (ruling out
a between-class aleatoric explanation), and (3) a unique sub-type (ruling out
within-class heterogeneity). This leaves the two epistemic quadrants as
candidates, and the quantities separate them cleanly: $\sigb$ ranks first of
ten classes and the class's retention (predict-to-train frequency ratio on
the model-unseen test split) is $0.000$: the class never wins. Its contested
reading is low ($\textrm{MI}=0.084$ on ``other''--ARJ21, the only pair in
which ARJ21 appears as a competitor, via top-2; its other pairs are
one-sided, \cref{sec:metrics}), as expected for a class that has
dropped out of competition rather than one fighting an under-trained boundary
(Appendix~\ref{sec:mi-blind}). The diagnosis is therefore unambiguous: ARJ21's
confusion is \textbf{collapsed epistemic}, which the framework labels
reducible. The predicted remedy is equally specific: supply more examples of
this class, and its recognition should recover while the irreducible
(geometric) sources stay put. We verify this prediction directly in
\cref{sec:treatment}.

\subsection{A220--A350: contested, reducible and supported}
\label{sec:case-a220a350}

A220 and A350 are not confused because they look alike on the ground. Their
physical size distributions barely overlap ($\textrm{BC}\approx0$):
narrow-body vs.\ wide-body. Yet this pair carries one of the two highest contested
readings among aircraft pairs ($\textrm{MI}=0.387$), the signature of a
boundary the model has not learned cleanly despite both classes being well
populated. The diagnosis is \textbf{contested epistemic}: a reducible,
under-trained boundary rather than an intrinsic overlap. The verdict is that
this confusion is improvable: targeted data or training aimed at this
boundary should reduce MI while leaving BC unchanged, since BC is a property
of the airframes, not of the training set.

Unlike the two aleatoric cases, this prediction is testable by intervention,
and we test it directly (\cref{sec:treatment-t2}): targeting the boundary
instances on which the ensemble most disagrees drives the pair's MI down by
more than half while its BC, a property of the airframes, stays fixed: the
signature of a boundary that was under-trained rather than intrinsically
overlapping.

\subsection{A330--Boeing787: affinity, irreducible by geometry}
\label{sec:case-a330b787}

A330 and Boeing787 are two wide-body airframes of nearly identical size. Their
physical size distributions overlap at $\textrm{BC}=0.945$, the second highest
of all pairs, while their contested uncertainty is low
($\textrm{MI}=0.047$). The diagnosis is \textbf{affinity aleatoric}: the two types
are close to indistinguishable by size geometry rather than facing an
under-trained boundary. The verdict is the opposite of the previous two cases:
since BC is a property of the airframes, it is unchanged by any amount of
training data, so this confusion is irreducible at the level of size geometry.
This verdict requires \emph{no} experiment: the framework states, from input
geometry alone, that data will not help, steering effort away from a wasted
remedy. Separating A330 from Boeing787 requires a different kind of
information (finer appearance features beyond OBB size), not more of the same.

\subsection{A330: heterogeneity, irreducible for a different reason}
\label{sec:case-hetero}

The same class, A330, also illustrates the fourth quadrant, and shows why
attribution matters, because its heterogeneity verdict is irreducible like
affinity yet calls for the opposite action. A330 covers two real sub-variants
of clearly different fuselage length (A330-200 \vs{} -300), which shows up in
its class-conditional size dispersion: body-length CV exceeds the noise floor
by $+0.011$ (floor $\approx0.062$), while wingspan CV sits at the floor:
exactly the signature expected when variants differ in length but share a
wingspan (\cref{fig:cv}, Appendix~\ref{app:hetero}). Three other multi-variant classes
behave the same way (Boeing737 body $+0.018$, A220 body $+0.009$,
Boeing787 body $+0.006$), while single-variant A321 sits at the floor. The attributed source is within-class
heterogeneity.

This heterogeneity has a measurable footprint on \emph{precision}: across the
non-collapsed classes, body-length excess-CV correlates negatively with
per-class precision (Spearman $\rho=-0.75$, $p=0.052$). The mechanism is
intuitive: a class with large intra-class size spread occupies a wider
region of size space, so neighboring-class instances fall inside it and are
mislabeled as it (precision down). A330 is the exemplar: recall $0.733$ but
precision only $0.375$. We report this as suggestive rather than causal, since
excess-CV also correlates with training count ($\rho=0.71$) in this small
sample.

The remedy verdict is the genuinely new one. Heterogeneity is aleatoric, so
like affinity it is irreducible, but for a different reason and with a
different consequence. Adding data does \emph{not} help: more examples of A330
do not shrink the intrinsic spread between its sub-variants (unlike the
collapsed case, where data is exactly the fix). The lever the
framework points to is to \emph{re-label} rather than \emph{re-sample}: split
the class into its sub-variants (A330-200, A330-300) so each sub-class is
geometrically tighter. An initial test did not confirm a gain: splitting A330
at a spec-anchored body-length threshold of $62.34$\,m (derived a priori from
the published fuselage lengths plus the annotation bias measured on A321),
with a matched split of single-variant A321 as control, changed AP by a
difference-in-differences of $+0.0001$, against a three-seed baseline AP range
of $0.025$. The null is bounded rather than decisive: at this resolution size
geometry alone misassigns $\sim\!26\%$ of instances even under a
Bayes-optimal split, and the minority sub-class falls into the scarcity regime
of \cref{sec:case-arj21}. We therefore state re-labeling as the distinctive
fourth verdict, one no other quadrant produces, but as a prediction not yet
confirmed on this dataset.

\subsection{A causal treatment: oversampling a collapsed class}
\label{sec:treatment}

The other three verdicts above are predictions. For the collapsed case we test the
prediction directly, with a single controlled intervention (denoted T1).

\paragraph{Design.}
If ARJ21's confusion is collapsed, i.e., data scarcity, not geometry, then
adding ARJ21 examples should reduce its collapsed quantities specifically,
without changing the aleatoric overlap of unrelated pairs. Specificity is the
point: we must distinguish ``repaired because we addressed the right cause''
from ``the model simply got better overall''. We oversample the $27$ training
images containing ARJ21 to $\sim\!10\times$ its original per-epoch occurrence,
concatenated with the full base set. Model, optimizer, schedule, checkpoint
selection (by overall mAP), and evaluation match the baseline; ARJ21 gets no
other special treatment. Since oversampling is at the image level, it also
repeats co-occurring classes; we quantify this by-catch in advance rather than
hide it.

\paragraph{Result.}
The diagnosis holds, and all four pre-registered predictions are confirmed
(\cref{tab:ap-t1}). ARJ21's val AP rises from $0.000$
to $0.333$ and its test AP from $0.000$ to $0.139$: strong, and still
confirmed on the harder, unseen split. Retention rises from $0.000$ to
$0.072$: the class re-enters the competition it had dropped out of. $\sigb$
falls by $12.5\%$, the \emph{only} one of ten classes to fall; the other
nine, including heavily by-caught ones, rise $2$--$23\%$ as the posterior
widens under continued training. ARJ21 narrowing \emph{against} this trend is
the signature of a targeted effect: we read $\sigb$ by direction relative to
the other classes, not by magnitude.

\paragraph{Two caveats.}
Two side effects are worth flagging rather than folding into the headline
result. First, A220's AP moves in \emph{opposite} directions on the two
splits ($+0.061$ val, $-0.045$ test); since A220 is not the treated class,
this is consistent with the val-side gain being partly a checkpoint-selection
artifact rather than a generalizing effect, so we do not count it either way.
Second, the $\sigb$ rise across the other nine classes above is a global
effect of the enlarged training set (the Laplace posterior is refit for the
whole model, not just ARJ21), not evidence that the intervention harms other
classes; it is the background trend against which ARJ21's own $\sigb$
\emph{fall} stands out as targeted.

\paragraph{Specificity: a natural control and the aleatoric side.}
Boeing777 is a control we did not have to design: also collapsed (retention
$0.000$ at baseline) but outside the oversampled images, so untreated. After T1
its retention is still $0.000$ and its $\sigb$ has \emph{risen} by $20.5\%$
with the general trend: generic improvement does not revive an untreated
collapsed class, so revival is targeted. (This natural control substitutes for
a designed placebo arm, left as confirmatory future work.) On the aleatoric
side, geometrically clean pairs (A330--Boeing787, A350--Boeing747) show
$C_{\textrm{ale}}$ drifting down uniformly ($\approx-0.04$), equally on clean
and contaminated pairs, a by-product of overall training progress (val mAP
$0.557\rightarrow0.622$), not an ARJ21-specific effect: no differential
aleatoric response, while BC is unchanged by construction. A targeted
collapsed quantity that moves while the aleatoric quantity does not is
exactly the dissociation the framework predicts.

\paragraph{From collapsed to contested.}
One further observation closes the loop with \cref{sec:case-a220a350}. After
treatment, eight of ARJ21's nine pairs move from the one-sided or
top-2-recovered populations into the valid tier: as the class stops
collapsing it re-enters the boundary competition MI can see, and its MI rises
accordingly (\eg{} ``other''--ARJ21 $0.084\rightarrow0.101$; A330--ARJ21
reaches $0.603$, though its one-sided baseline value is not directly
comparable). The class moves from collapsed (MI-blind) into contested
(MI-visible), direct dynamic evidence that the two epistemic sources are distinct and that the
collapsed one is reducible.

\begin{table}[t]
\centering
\caption{Verification of T1 (ARJ21 collapsed-epistemic intervention):
pre-registered predictions vs.\ observed outcomes. All four predictions on
ARJ21 are confirmed; the aleatoric-reference rows show no matching targeted
response. The untreated collapsed control (Boeing777) is discussed in the
text. See \cref{sec:treatment}.}
\label{tab:ap-t1}
\begin{tabular}{lccl}
\toprule
\textbf{Metric} & \textbf{Predicted} & \textbf{Observed} & \textbf{Verdict} \\
\midrule
val AP           & $\uparrow$ & $0.000 \rightarrow 0.333$ & $\checkmark$ strong \\
test AP          & $\uparrow$ & $0.000 \rightarrow 0.139$ & $\checkmark$ (test harder than val) \\
$\sigb$          & $\downarrow$ & $-12.5\%$ & $\checkmark$ targeted narrowing \\
retention (test) & $\uparrow$ & $0.000 \rightarrow 0.072$ & $\checkmark$ targeted revival \\
$C_{\textrm{ale}}$ ref. & no change & $\approx-0.04$ (uniform) & no targeted response \\
BC               & no change & unchanged & geometry is data-independent \\
\bottomrule
\end{tabular}
\end{table}

\subsection{A second treatment: sharpening a contested boundary}
\label{sec:treatment-t2}

If A220--A350 is contested, i.e., an under-trained but learnable boundary rather
than an intrinsic overlap, then more exposure to the instances the model is
most unsure about should sharpen the boundary and lower the pair's MI while
leaving BC untouched. We test this (T2) by oversampling, at the same
$10\times$ rate as T1, the fifty development images richest in
high-disagreement A220/A350 instances (ranked by per-instance ensemble
standard deviation), under the otherwise-identical protocol.

The main effect holds and is confirmed on both splits (\cref{tab:ap-t2}). The
A220--A350 MI falls from $0.387$ to $0.183$ ($-53\%$), and the ensemble
disagreement on the treated boundary instances narrows accordingly (mean
member-std $0.24\rightarrow0.15$; fraction with std $>0.2$ drops from
$61\%$ to $41\%$). A350's AP improves on both val ($0.698\rightarrow0.788$)
and test ($0.582\rightarrow0.619$), and BC is unchanged by construction, so
the confusion was reduced without touching the geometry: consistent with the
contested verdict.

\begin{table}[t]
\centering
\caption{Verification of T2 (A220--A350 contested-epistemic intervention):
pre-registered predictions vs.\ observed outcomes. The main effect (MI drop,
A350 improvement, BC unchanged) is confirmed on both splits; A220's
val/test divergence was not predicted and is a caveat discussed in the
text.}
\label{tab:ap-t2}
\begin{tabular}{lccl}
\toprule
\textbf{Metric} & \textbf{Predicted} & \textbf{Observed} & \textbf{Verdict} \\
\midrule
A220--A350 MI & $\downarrow$ & $0.387\rightarrow0.183$ ($-53\%$) & $\checkmark$ strong \\
A350 val AP   & $\uparrow$ & $0.698\rightarrow0.788$ ($+0.090$) & $\checkmark$ \\
A350 test AP  & $\uparrow$ & $0.582\rightarrow0.619$ ($+0.037$) & $\checkmark$ \\
BC (overlap)  & unchanged & replicated from baseline & $\checkmark$ by construction \\
A220 val AP   & --- (not predicted) & $+0.027$ & neutral-to-positive \\
A220 test AP  & --- (not predicted) & $-0.041$ & opposite sign from val \\
\bottomrule
\end{tabular}
\end{table}

\paragraph{A caveat: a val/test divergence on A220.}
A220 was not the treated class, and its AP moves in \emph{opposite}
directions on the two splits: $+0.027$ on val, $-0.041$ on test. This
divergence is a more informative signal than either number alone. If the
intervention had taught the model a genuinely better A220/A350 boundary,
A220's own AP should not fall on the held-out split. The pattern is instead
consistent with the model resolving ensemble disagreement (MI$\downarrow$) by
collapsing toward a single, A350-leaning prediction rather than by learning
new discriminative information: on val, this shift happens to land on
instances where it still counts as correct (apparent AP gain); on test, the
same shift costs A220 detections that would otherwise have counted correctly
(the loss surfaces as an AP drop). The retention numbers point the same
way: A220's over-prediction eases from $1.59$ (baseline) to $1.19$ (under
T2), and A350's under-prediction recovers from $0.50$ to $0.73$, a shift of
predictions \emph{away from} A220 and
\emph{toward} A350, which is compatible with either a genuinely sharpened
boundary or this collapse-toward-A350 reading.
We therefore treat the main effect as confirmed but not yet disentangled from
this alternative mechanism.

We also flag that this is a directional confirmation rather than a fully
isolated effect. Continued training sharpens the posterior globally (all
$\sigb$ drop $\sim\!70\%$, val mAP $+0.03$), so part of the MI decrease is
global; the A220--A350 drop is nonetheless the eighth-largest of all
forty-five pairs and well beyond the average pairwise change: a global
sharpening plus a clear pair-specific excess. Because oversampling is at the
image level, it also inflates the A220/A350 populations as a whole, so
exposure and targeting are entangled; a fully specific test, isolating
boundary instances without class-level inflation and separating genuine
boundary learning from prediction collapse, is left to future work.

\section{Limits of Measurement}
\label{sec:cannot}

A source-attribution framework is only honest if it states where its
instruments run out. The decomposition (\cref{sec:framework}) enumerates the
\emph{types} of confusion source, but this does not make all four equally
\emph{measurable} on a given dataset. We report two such limits:
the heterogeneity quadrant is observable but noise-limited, and the collapsed
quantity is well defined but directionless. In both we give the graded
statement rather than overclaim.

\subsection{Heterogeneity is partially observable, and noise-limited}
\label{sec:cannot-within}

The within-class aleatoric quadrant is not empty here, but its signal is
partial. Conceptually it has two sub-components: (1) sub-variant geometric
diversity: different airframes sharing one label (\eg{} A330-200 \vs{}
-300); and (2) imaging-level appearance dispersion (blur, illumination,
compression). We measure the first and place the second out of scope.

\paragraph{What is observable.}
Using the excess-CV metric of \cref{sec:metrics}, four multi-variant,
non-collapsed classes carry body-length dispersion above the noise floor
(Boeing737 $+0.018$, A330 $+0.011$, A220 $+0.009$, Boeing787 $+0.006$), while single-variant
reference A321 sits at the floor and A350, single-variant here, its
dominant type covering $\sim\!90\%$ of instances, sits below it
(\cref{fig:cv}, Appendix~\ref{app:hetero}). The above-floor classes are exactly those
with known fuselage-length sub-variants, consistent with sub-variant
heterogeneity, which is what revived the quadrant.

\paragraph{Where it runs out.}
The signal is real but weak: measurement noise is of the same order as the
variant spacing (the single-variant floor of CV $\approx0.06$ is about
$\pm3.7$\,m on a $61$\,m fuselage, while the A330-200/-300 spacing is only
$\sim\!5$\,m). No statistic separates the classes cleanly: excess-CV, a
Hartigan dip test, and kernel-density peak counting give mutually inconsistent
orderings (the dip test even ranks single-variant A321 above double-variant
A330), and a raw bimodality coefficient stays below its $0.555$ threshold
everywhere, since the sub-variant modes are unequal and merge
(Appendix~\ref{app:hetero}). The statement is therefore graded: classes above the
floor are \emph{consistent with} heterogeneity; classes at or below it are
\emph{inconclusive}, i.e., insufficient evidence, not evidence of absence. The
quadrant is observable but not identifiable at the level of clean anchors,
since OBB resolution is comparable to the variant geometry it would resolve.

\paragraph{Imaging-level dispersion is out of scope, not a gap.}
The second sub-component, dispersion from blur, illumination, and
compression, is deliberately out of scope: a data-quality factor applied
roughly uniformly across classes, it lowers overall accuracy but does not by
itself generate the class-specific structure of a confusion (it does not
explain why class $A$ is taken for $B$ rather than $C$), so it is orthogonal
to the source decomposition, not a fifth quadrant. It can be quantified
(\eg{} by test-time-augmentation variance), but its remedy lies in sensor
choice or acquisition, outside the model- and training-level actions
\ensuremath{A^2E^2} is built to guide.

\subsection{The collapsed quantity has no direction}
\label{sec:cannot-direction}

The collapsed quantity $\sigb$ is the posterior width of a class's bias, a
per-class scalar. It can report \emph{that} a class is collapsing, but not
\emph{which} competitor it collapses toward: direction is a two-class relation
living in the cross-class covariance of the last-layer posterior, which the
diagonal Laplace approximation sets to zero by construction. Recovering it
would need a more expensive, harder-to-stabilize off-diagonal posterior. We
regard the undirected form as the instrument's honest scope and leave a
directed treatment to future work.

\section{Conclusion}
\label{sec:conclusion}

We have presented \ensuremath{A^2E^2}, a source-attribution framework decomposing
confusion sources along two dimensions,
$\{\textrm{aleatoric}, \textrm{epistemic}\} \times \{\textrm{within-class},
\textrm{between-class}\}$. It turns a confusion matrix, which records
\emph{where} a model is confused, into a diagnosis with a cause and a remedy
verdict. (1) Two
exhaustive binary partitions enumerate the source types. (2) Each quadrant has
a single quantity, computed in the input geometry, the output-space ensemble,
or the bias-parameter posterior; the partition is structural, while the
quantities are complementary readings rather than independent ones. (3) The epistemic side is two
phenomena: a contested boundary MI can see, and a collapsed class MI is
blind to but $\sigb$ captures. (4) The four verdicts differ: collapsed is
repaired by targeted data (verified); contested is improvable by boundary
data; affinity is irreducible from size alone, flagged without any
experiment; heterogeneity is also irreducible by data and points to
re-labeling into sub-variants, a prediction an initial test could not yet
confirm; these are opposite actions no single confusion count could distinguish.

We are also explicit about the instrument's limits. The heterogeneity quadrant
is observable but noise-limited: several multi-variant classes carry size
dispersion above the floor, consistent with their sub-variants, but the OBB
noise is of the same order as the variant spacing, so the reading is graded
(consistent-with / inconclusive) rather than a clean per-class identification.
The collapsed quantity, under a diagonal posterior, reports that a class
collapses but not toward which competitor. Neither is a failure of the
taxonomy; stating them plainly is what lets us claim the rest with confidence.
For future work, we would sharpen the two partial quantities (an
off-diagonal posterior for collapse direction, a higher-resolution geometric
measure for heterogeneity) and validate the framework across further
datasets and detectors; the aleatoric axis relies on oriented-box geometry,
informative for aircraft but needing a different measure elsewhere. The
contested boundary-sharpening intervention confirms its prediction
directionally; a fully specificity-isolated version and a better-powered
re-labeling test for heterogeneity would close the last open loops. Grounding confusion in
measured, attributable sources is, we believe, a useful step toward
trustworthy fine-grained detection.

\bibliographystyle{splncs04}
\bibliography{main}

\newpage
\appendix

\section{Completeness over Source Types}
\label{app:completeness}
This appendix details the claim that the \ensuremath{A^2E^2} quadrants enumerate the source types (\cref{sec:quadrants}), which rests on
two observations. First, the aleatoric/epistemic split is exhaustive. We adopt
the standard dichotomy~\cite{kendall2017uncertainties, huellermeier2021aleatoric}:
any predictive uncertainty is either aleatoric (intrinsic and irreducible) or
epistemic (from limited knowledge and reducible). We treat distributional or
out-of-distribution uncertainty as a sub-case of the epistemic term (the
model's ignorance of unseen regions), not as a separate axis, consistent with
the prevailing formulation. Second, the within/between split is exhaustive by
construction. Confusion is a class-level event: any single confusion occurs
either within a class (one intra-class variant taken for another) or across a
class boundary (one class taken for another). This is a mutually exclusive,
exhaustive partition and needs no empirical validation. Degenerate classes do
not break it: assigning an aircraft to ``other'' or to background is still
a between-class crossing.

The Cartesian product of two exhaustive binary dimensions is itself
exhaustive, so the four quadrants cover all source types. We claim completeness
of the \emph{types}; measurability of each in a given dataset is a separate,
empirical question. All four quadrants are populated here: the within-class
aleatoric (heterogeneity) quadrant, in particular, is not empty: several
multi-variant classes show intra-class size dispersion above the measurement
noise floor, consistent with their sub-variant geometry (\cref{sec:metrics},
\cref{sec:cannot}). Its signal is partial and noise-limited rather than absent,
and we report both what it reveals and where it runs out.

\section{Physical Size Geometry of the Ten Classes}
\label{app:sizegeo}

This appendix gives the visual counterpart to two claims made in the main
text: the affinity/contested contrast of \cref{sec:exp}, where A330--Boeing787
and A220--A350 sit at opposite corners of the framework precisely because of
their physical size geometry, and the aleatoric-geometry correlation reported
in Appendix~\ref{app:orthogonality}. \Cref{fig:scatter} plots each class's body length
against wingspan directly, making the BC-based overlap and separation of
\cref{eq:bc} visible on the raw physical measurements.

\begin{figure}[h!]
\centering
\includegraphics[width=0.78\textwidth]{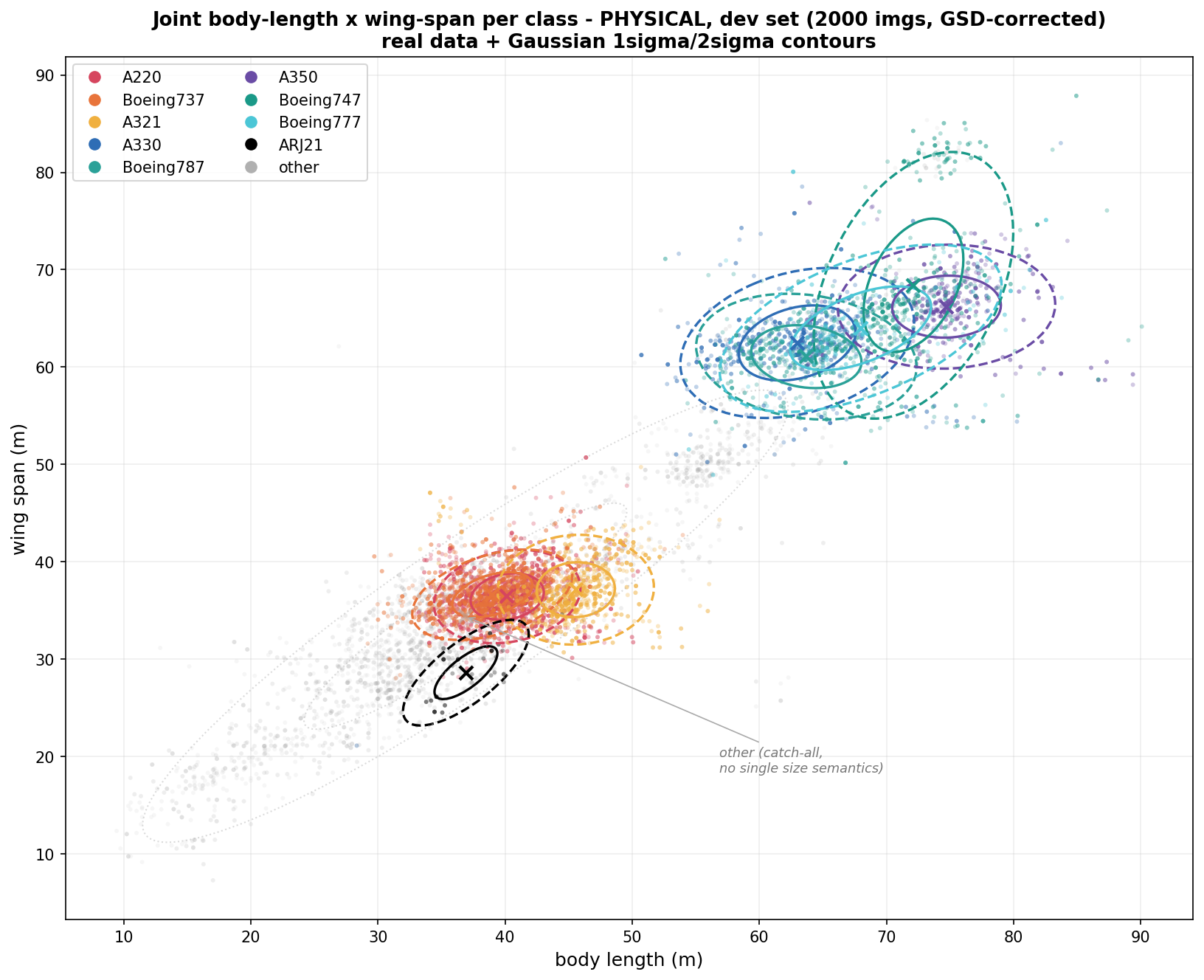}
\caption{Per-class physical size geometry (body length vs.\ wingspan, meters,
GSD-corrected) on the development set, with Gaussian $1\sigma/2\sigma$
contours. Narrow- and wide-body aircraft are well separated (BC $\approx 0$).}
\label{fig:scatter}
\end{figure}

\section{Separation of the Four Quantities}
\label{app:orthogonality}

A decomposition is only useful if its sources can be told apart: if the
quantities measured a single underlying thing, attributing a confusion to one
source rather than another would be arbitrary. We argue that the partition of
\ensuremath{A^2E^2} is \emph{structural} and that its quantities read different
parts of the pipeline, and we then support this empirically, while being
explicit about where the evidence is strong and where it is weak. We do not
claim that the four quantities are statistically independent: the two
aleatoric quantities share their input (Appendix~\ref{sec:ortho-structural}).

\subsection{The quantities reside in different parts of the model}
\label{sec:ortho-structural}

The four quantities defined in \cref{sec:metrics} are computed in three
places:

\begin{itemize}
\item BC (between-class aleatoric) and excess-CV (within-class aleatoric) are
functions of the \emph{input geometry} only. They are computed from physical
OBB sizes and never read the model's weights or predictions; both would be
unchanged if the model were retrained from scratch.
\item MI (contested epistemic) is a function of the \emph{output space}. It is
the disagreement among posterior samples about which of two classes wins, read
from the renormalized two-class predictive distribution.
\item $\sigb$ (collapsed epistemic) is a function of the \emph{bias-parameter
posterior}. It is the posterior width of a single scalar parameter, the class
bias, and is input-independent by construction (\cref{eq:sigb}).
\end{itemize}

Because a size overlap, an output-space disagreement, and a bias-parameter
variance are computed from disjoint parts of the system, no algebraic identity
forces them to move together: in this sense the separation between the two
epistemic quantities, and between each of them and the input geometry, holds by
construction rather than as a low correlation on this dataset. The two
aleatoric quantities are the exception. BC depends on each class's size
covariance $\Sigma_a$, whose diagonal is the variance underlying CV, so a class
with larger dispersion also tends to overlap more with its neighbors.
Empirically the coupling is weak here (Pearson $r=+0.36$ between excess body
CV and a class's maximum BC, $n=9$), because inter-class distance dominates
BC; but weak correlation is not independence, and we treat the two as
complementary readings of the same geometry. The two epistemic quantities deserve particular
emphasis, because they might be expected to coincide. They do not: MI lives in
the output space and measures disagreement \emph{between} two classes, while
$\sigb$ lives in the bias term and measures a single class's prior tendency.
They can therefore diverge for the same class, and the next subsection shows
that one of them is in fact blind to a phenomenon the other captures.

\subsection{Why MI is blind to a collapsed class}
\label{sec:mi-blind}

MI measures whether the posterior ensemble \emph{disagrees} about which of two
classes wins. Consider a class with very few training examples that has
dropped out of competition: across posterior samples it consistently loses to
its competitor by a stable margin. The ensemble \emph{agrees} on the outcome
(the rare class loses), even though that agreement is itself a symptom of data
starvation. MI is the mutual information between prediction and parameters, so
when the ensemble agrees, MI is near zero: precisely when the class is most
data-starved. This is a general property of the BALD decomposition, not an
artifact of our implementation: disagreement-based epistemic measures read
\emph{low} for a class that is stably, confidently losing. Toure and
Stephens~\cite{toure2026notjust} formalize the same suppression for per-class
decompositions of MI: the per-class predictive variance is bounded by
$\mu_k(1-\mu_k)$, and even their rescaled per-class contribution vanishes for a
class whose mean predicted probability is zero. $\sigb$, which lives in
parameter space, does not require the class ever to be predicted.

This is the structural reason the epistemic side needs two quantities rather
than one. A contested boundary produces disagreement and is caught by MI; a
collapsed class produces stable agreement and is missed by MI, but is caught by
the bias-posterior width $\sigb$, which reflects how little evidence has
constrained the class's prior. Merging the two into a single epistemic scalar
would discard exactly this distinction. The collapsed phenomenon lives in the
bias parameter and the contested phenomenon lives in the output prediction, so
no single output-space scalar can represent both.

\subsection{Empirical support, reported by strength}
\label{sec:ortho-empirical}

The structural separation is an argument about construction; we now ask what the
data show. We separate strong evidence (relationships with adequate sample
size, and a clean single-class contrast) from weak evidence (a per-class
correlation that is underpowered), and we report the weak evidence as such.

\paragraph{The aleatoric quantity tracks geometry, and a control runs the
other way.}
Across pairs, $C_{\textrm{ale}}$ from \cref{eq:mi} correlates with the
physical size overlap BC at Spearman $\rho=+0.746$ (aircraft-only, $n=35$,
$p<10^{-4}$), strengthening to $\rho=+0.825$ when the ``other'' category
is included ($n=44$). As a control, $C_{\textrm{ale}}$ correlates with the
centroid distance between class size distributions in the opposite direction
($\rho\approx-0.65$): aleatoric confusion rises as airframes overlap in size
and falls as they separate. The two directions match the prediction, and the
relationship is computed on enough pairs to be meaningful.

\paragraph{The contested quantity tracks data scarcity.}
Per-pair MI correlates with the smaller of the two classes' training counts at
$\rho=-0.688$ ($n=28$ valid pairs, $p\approx5\times10^{-5}$): the largest
contested uncertainty clusters at the smallest training counts, as expected for
a reducible, under-trained boundary.

\paragraph{The collapsed quantity is cross-validated at an anchor, not by a
per-class correlation.}
The cleanest evidence that collapsed epistemic is a distinct source is a
single-class contrast on ARJ21, the rarest aircraft class, cross-validated by
two measures computed along different paths (\cref{tab:arj21}). ARJ21 reads
collapsed-extreme on a parameter-side measure ($\sigb$ ranks first of ten
classes) \emph{and} on a count-side measure (a retention ratio of raw
predict-to-train frequency of $0.000$), while its contested reading is only
low-to-moderate (MI $=0.084$ for ``other''--ARJ21, the only pair in which
ARJ21 competes, recovered via top-2 with $n=300$). One quantity is derived from
the model's Laplace posterior and the other from prediction counts on the test
split; both are ultimately driven by the class's scarcity (\cref{sec:metrics}),
so their agreement confirms the collapsed reading rather than establishing it
independently. What separates the collapsed from the contested source is the
low MI on the same class.

\begin{table}[t]
\centering
\caption{Single-class cross-validation on ARJ21, the rarest class. Two
measures computed along different paths (Laplace posterior and prediction
counts) both read collapsed-extreme, while the output-space contested measure
reads low; see Appendix~\ref{sec:ortho-empirical}.}
\label{tab:arj21}
\begin{tabular}{lll}
\toprule
\textbf{Measure (source)} & \textbf{ARJ21} & \textbf{Reading} \\
\midrule
$\sigb$ (collapsed, Laplace bias posterior) & $6.48\times10^{-3}$ & rank 1 of 10 \\
retention ratio (collapsed, raw counts)     & $0.000$            & tied-lowest \\
MI (``other''--ARJ21) (contested, output space) & $0.084$    & low-to-moderate \\
\bottomrule
\end{tabular}
\end{table}

\paragraph{The per-class correlation is weak, and we do not lean on it.}
A direct per-class Spearman between the contested and collapsed readings is
weak and non-significant ($\rho=-0.273$, $p=0.45$), which is consistent with
orthogonality but is not strong evidence for it. We flag this explicitly as
underpowered: with only ten classes there is little power to separate ``truly
orthogonal'' from ``weakly correlated'', and the contested reading for the
anchor classes rests on very few measurable pairs (one for ARJ21). We therefore
treat the per-class correlation as supporting, not primary, evidence, and rest
the separation claim on the structural argument
(Appendix~\ref{sec:ortho-structural}, Appendix~\ref{sec:mi-blind}) and the anchor
cross-validation above.

\section{Rejection of the Weight-Side Variance $\sigW$}
\label{app:sigmaW}

A tempting last-layer candidate for within-class heterogeneity is the
weight-side posterior variance $\sigW\equiv\sigma^2_{W,X,d}$ from
\cref{eq:sigb}, the input-dependent part of a class's logit variance. It is
rejected because it does not measure sub-variant geometry. Two tests show this.
First, $\sigW$ does not track a direct sub-variant probe: its correlation with
the within-class size dispersion of the multi-variant classes is weak and not
significant ($\rho\approx+0.4$). Second, $\sigW$ correlates strongly and
negatively with training-sample count ($\rho=-0.91$), the same quantity that
drives the collapsed-epistemic term. In other words, the only clean signal in
the weight-side variance is data scarcity, not sub-variant geometry: $\sigW$ is
collapsed-epistemic in disguise. What is rejected is the \emph{metric}, not the
\emph{concept}; the within-class aleatoric quadrant remains valid and is
measured instead by the input-geometry excess-CV of \cref{sec:metrics}.

\section{Rejected Heterogeneity Metrics}
\label{app:hetero}

\begin{figure}[t]
\centering
\includegraphics[width=\textwidth]{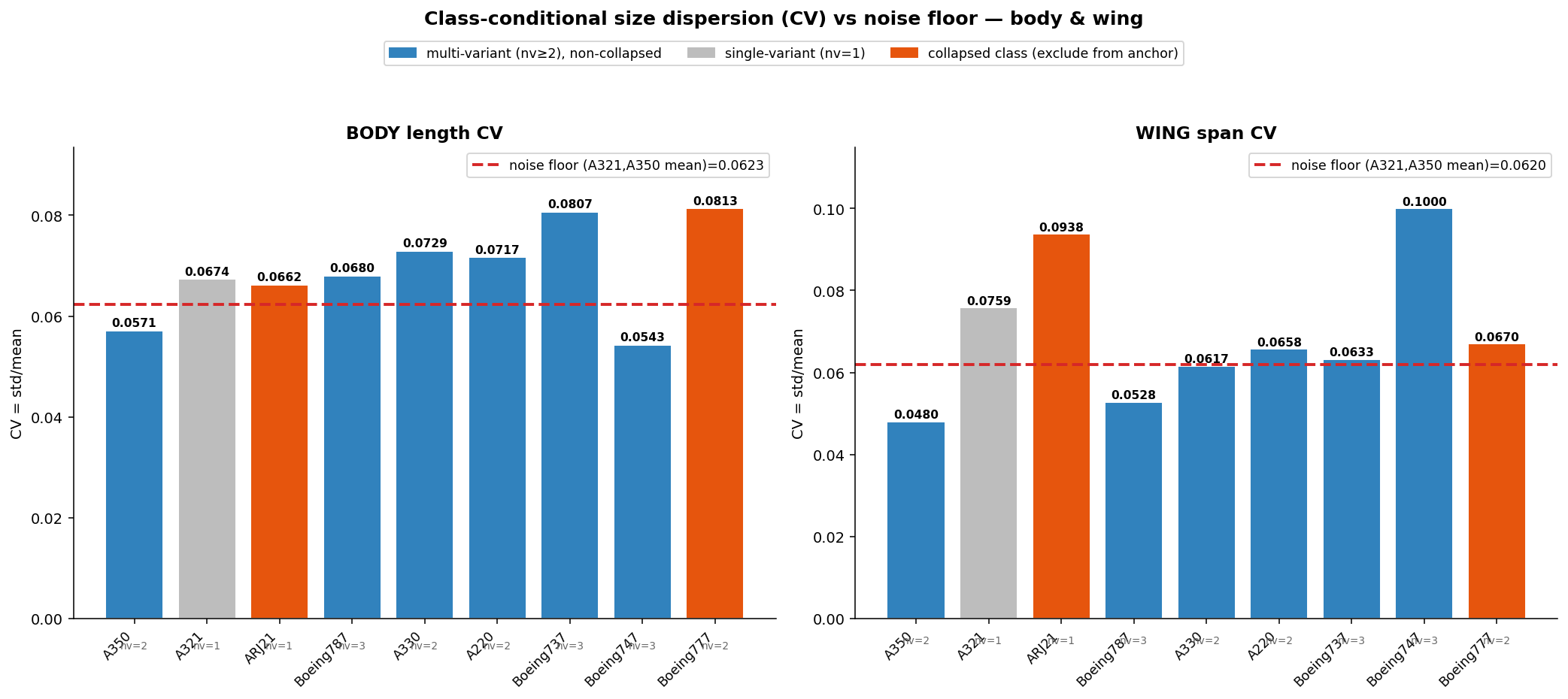}
\caption{Heterogeneity metric: class-conditional size CV against the noise
floor (dashed, mean CV of the reference classes A321/A350, $\approx0.062$;
A350 is drawn as multi-variant by catalogue, but its instances here are
predominantly one variant, \cref{sec:metrics}).
\textbf{Left}: body length; the multi-variant classes B737/A330/A220/B787
exceed the floor, consistent with their sub-variants; single-variant A321
(gray) sits at it. \textbf{Right}: wingspan; the same classes sit near the floor, since
their variants share a wingspan. Collapsed classes (orange, ARJ21/B777) are
excluded from anchoring: their dispersion is confounded by scarcity and, for
B747/B777 wingspan, by OBB artifacts. See \cref{sec:cannot-within}.}
\label{fig:cv}
\end{figure}

\begin{figure}[h!]
\centering
\includegraphics[width=0.7\textwidth]{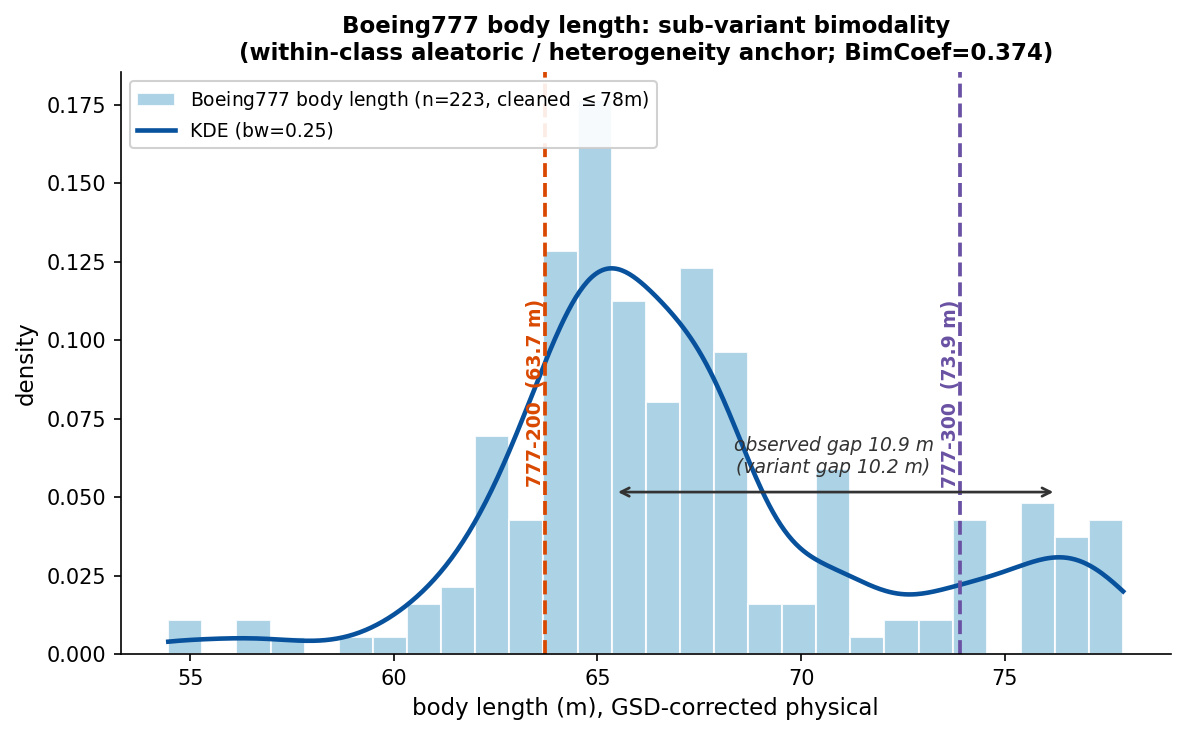}
\caption{Boeing777 body length shows a genuine bimodal structure whose peak
spacing ($\approx11$\,m) matches the 777-200/-300 fuselage difference
($10.2$\,m). However, Boeing777 is a collapsed class, so this within-class
signal is confounded with data scarcity and we do not use it as a heterogeneity
anchor; it is shown here as a rejected/contaminated case.}
\label{fig:b777}
\end{figure}

Because the OBB measurement noise is of the same order as the sub-variant
spacing (\cref{sec:cannot-within}), several natural bimodality metrics fail to
order the classes correctly, which is why we adopt the graded excess-CV reading
instead (\cref{fig:cv}). (i) \emph{Bimodality coefficient} (Sarle): stays below the $0.555$
threshold for every class, because it is derived from skewness and kurtosis and
is insensitive to the unequal, merged sub-variant modes seen here: it reads
the clearly bimodal A330 body at only $0.14$. (ii) \emph{Hartigan dip test}:
gives inconsistent significance, ranking single-variant A321 above
double-variant A330, because the dip is comparable to noise. (iii)
\emph{Peak finding}: the apparent second peak in Boeing747 wingspan sits at
$\approx82$\,m, above every real 747 variant, \ie{} it is an OBB artifact rather
than a sub-variant; Boeing777 body length is genuinely bimodal (two peaks at a
$\sim\!11$\,m spacing, matching the 777-200/-300 fuselage difference), but
Boeing777 is a collapsed class, so its within-class reading is confounded by
scarcity. \Cref{fig:b777} shows the Boeing777 case. None of these yields a
clean, dataset-wide ordering, whereas excess-CV over a single-variant floor at
least separates the multi-variant classes from the single-variant references in
the body dimension.

\end{document}